\documentclass[5p,authoryear]{elsarticle} 

\usepackage{lineno,hyperref}
\modulolinenumbers[5]
\usepackage{amsmath}
\usepackage{amssymb}
\usepackage{verbatim}
\usepackage{multirow}

\usepackage{color,graphicx}
\usepackage{arydshln}

\usepackage{hyperref}
\usepackage{url}
\usepackage{footnote}
\usepackage{multirow}
\usepackage{bm}

\usepackage{cite}
\usepackage{subfigure}

\providecommand{\Zxhreftb}[1]{Table~\ref{#1}}
\providecommand{\zxhreftb}[1]{Table~\ref{#1}}
\providecommand{\zxhreffig}[1]{Fig.~\ref{#1}}
\providecommand{\Zxhreffig}[1]{Fig.~\ref{#1}}

\journal{Medical Image Analysis}

\begin{document}

\begin{frontmatter}
\title{Atrial Scar Quantification via Multi-scale CNN in the Graph-cuts Framework}

\author[label1,label2]{Lei Li} 
\author[label3]{Fuping Wu} 
\author[label4,label5]{Guang Yang} 
\author[label6]{Lingchao~Xu} 
\author[label5]{Tom Wong} 
\author[label4,label5]{Raad Mohiaddin}  
\author[label4,label5]{David Firmin} 
\author[label4,label5]{Jennifer Keegan} 
\author[label2]{Xiahai Zhuang*} 
\ead[url]{zxh@fudan.edu.cn}

\address[label1]{School of Biomedical Engineering, Shanghai Jiao Tong University, Shanghai, China}
\address[label2]{School of Data Science, Fudan University, Shanghai, China}
\address[label3]{Dept of Statistics, School of Management, Fudan University, Shanghai, China}
\address[label4]{National Heart and Lung Institute, Imperial College London, London, UK}
\address[label5]{Cardiovascular Research Center, Royal Brompton Hospital, London, UK}
\address[label6]{School of NAOCE, Shanghai Jiao Tong University, Shanghai, China}

\begin{abstract}
Late gadolinium enhancement magnetic resonance imaging (LGE MRI) appears to be a promising alternative for scar assessment in patients with atrial fibrillation (AF).
Automating the quantification and analysis of atrial scars can be challenging due to the low image quality.
In this work, we propose a fully automated method based on the graph-cuts framework, where the potentials of the graph are learned on a surface mesh of the left atrium (LA) using a multi-scale convolutional neural network (MS-CNN).
For validation, we have employed fifty-eight images with manual delineations.
MS-CNN, which can efficiently incorporate both the local and global texture information of the images, has been shown to evidently improve the segmentation accuracy of the proposed graph-cuts based method.
The segmentation could be further improved when the contribution between the t-link and n-link weights of the graph is balanced.
The proposed method achieves a mean accuracy of $0.856 \pm 0.033$ and mean Dice score of $ 0.702 \pm 0.071 $ for LA scar quantification.
Compared with the conventional methods, which are based on the manual delineation of LA for initialization, our method is fully automatic and has demonstrated significantly better Dice score and accuracy ($p<0.01$).
The method is promising and can be useful in diagnosis and prognosis of AF.
\end{abstract}


\end{frontmatter}

\section{Introduction}
Atrial fibrillation (AF) is the most common arrhythmia observed in clinical practice, occurring in up to 1\% of the population and rising fast with advancing age \citep{journal/Circulation/chugh2013}.
Radiofrequency catheter ablation using the pulmonary vein (PV) isolation technique has emerged as one of the most common methods for the treatment of AF patients \citep{journal/jama/Wilber2010,journal/hr/Calkins2012}.
Quantification of atrial scars is potentially beneficial in selecting candidates and guiding ablation treatment.
Late gadolinium enhancement magnetic resonance imaging (LGE MRI) is a promising technique to visualize and quantify the atrial scars \citep{journal/jce/Vergara2011}.
Many clinical studies mainly focus on the location and extent of scarring areas of the left atrium (LA) myocardium \citep{journal/jacc/Mcgann2008,journal/hr/Vergara2011,journal/cae/Badger2010}.

\begin{figure*}[t]\center
    \subfigure[] {\includegraphics[width=0.32\textwidth]{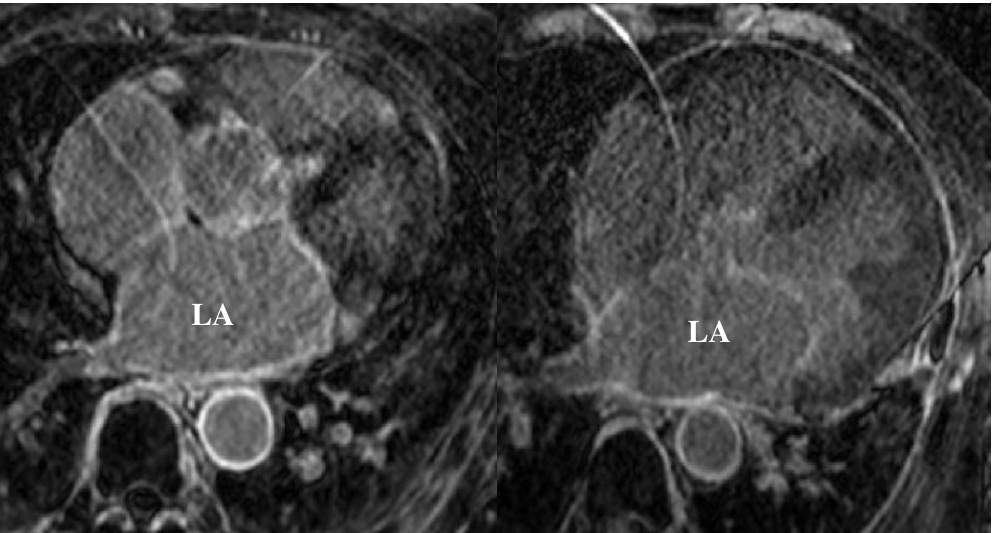}}
    \subfigure[] {\includegraphics[width=0.32\textwidth]{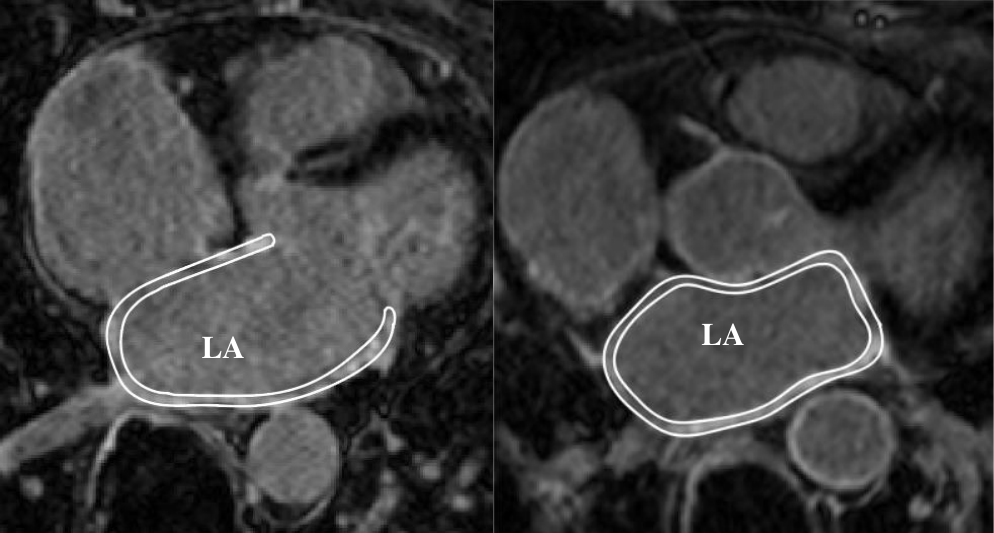}}
    \subfigure[] {\includegraphics[width=0.32\textwidth]{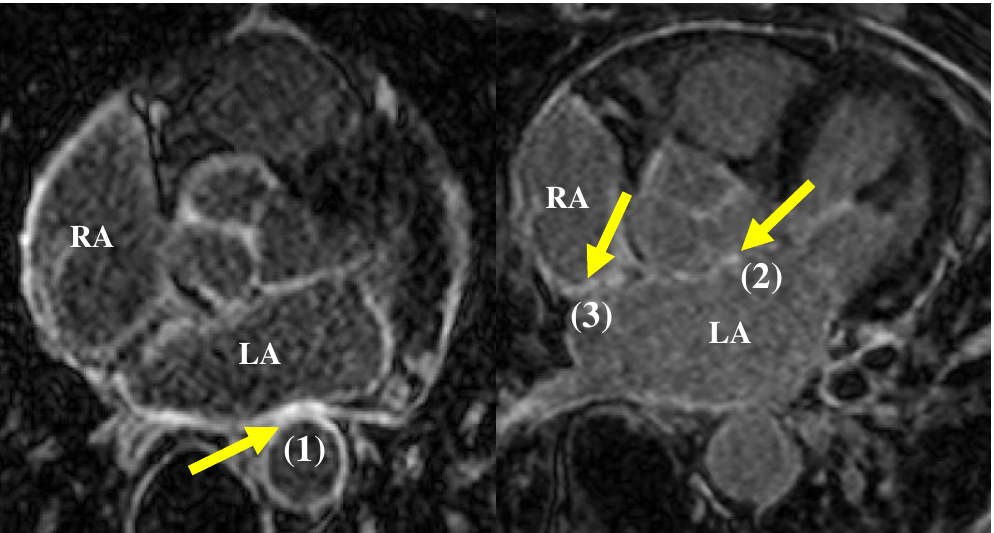}}
   \caption{The challenges of automatic delineation of scars from LGE MRI:
     (a) two typical LGE MRIs with poor quality; 
     (b) thin atrial walls highlighted using bright white color in the figure;
     (c) surrounding enhanced regions pointed out by the arrows, where (1) and (2) respectively indicate the enhanced walls of descending and ascending aorta, (3) denotes the enhanced walls of right atrium.}
\label{fig:challenges}\end{figure*}

Automatic delineation of scars from LGE MRI is still challenging due to various reasons.
First, the image quality of LGE MRI is generally poor.
Second, the prior model of scars is hard to construct on account of the various LA shapes, the thin wall (mean thickness of $ 1.89 \pm 0.48 $ mm reported by \citet{journal/jce/beinart2011}), the surrounding enhanced regions and the complex patterns of scars in AF patients.
\Zxhreffig{fig:challenges} illustrates and explains the challenges in more details.
To the best of our knowledge, little work has been reported in the literature to achieve the fully automatic quantification of LA scars from LGE MRI.

The most widespread methods for atrial scar segmentation are mainly based on thresholding \citep{journal/cae/Badger2010, journal/jcmr/Karim2013}.
\citet{journal/euo/Pontecorboli2016} provided an overall review of scar segmentation using various threshold techniques.
For these methods, an appropriate threshold value is decisive, but setting this value can be subjective, eventually limiting the applicability and reproducibility.
\citet{conf/mi/Perry2012} proposed to use k-means clustering to classify the normal and fibrosis tissue from manually segmented LA walls.
\citet{journal/jtehe/Karim2014} combined the scar intensity priors and Gaussian mixture model (GMM) to construct a cost function for scar segmentation, which was achieved by an optimization using the graph-cuts framework.
\citet{journal/mp/Yang2018} employed the super-pixel method and support vector machine (SVM) to segment the atrial scars.

Most of the reported methods rely on manual segmentation of the LA or LA walls to provide an accurate initialization.
In ISBI 2012 challenge \citep{journal/jcmr/Karim2013}, manual segmentation of LA was provided.
There was large variance in terms of segmentation accuracy, especially for the pre-ablation cases,
and the teams using manually delineated LA walls generally obtained much better performance than those using fully automatic approaches in the challenge.
Their benchmark study emphasizes the importance of an accurate initialization.

For LA segmentation, \citet{journal/tmi/Ravanelli2014} proposed a method using threshold for initialization, followed by the 3D fast marching for segmentation.
They required manual correction from the clinicians to achieve reliable performance.
\citet{journal/jmri/Tao2016} combined LGE MRI with another MRI sequence with better anatomical information to segment the LA, and the combined segmentation achieved better results than the method solely using LGE MRI.
\citet{journal/tmi/xiong2018} proposed a dual fully convolutional neural network for LA segmentation from LGE MRI with promising results.
Later, they organized a LA segmentation challenge in MICCAI 2018 \citep{link/LAseg2018}.
For LA wall segmentation, \citet{journal/mia/Veni2017} proposed an algorithm named ShapeCut, combining a shape-based system and the graph-cuts approach to make a Bayesian dual surface estimation.
\citet{conf/ip/Ji2018} applied the advanced two-layer level set with a soft distance constraint for dual surface segmentation of LA and LV walls.
Their method was 2D-based and required an manual initialization of the endocardial boundaries.

In summary, in previous works scar quantification relies on an accurate segmentation of the LA or LA walls for initialization,
but automating this segmentation is still an open question.
In this work, we propose a fully automatic method for LA scar quantification and analysis, without the requirement of an accurate LA segmentation.

Firstly, we propose to perform scar quantification on a surface, onto which the LA endocardium is projected.
We neglect the thickness of LA walls, because the clinical studies are generally performed by projecting the scars onto the LA endocardial surface for visualization \citep{journal/Radiology/Peters2007,journal/ibme/Knowles2010,journal/tmi/Ravanelli2014}.
In this framework, we represent the surface using a graph, and formulate the classification as an energy minimization problem which can be solved by graph-cuts.
We further propose to explicitly learn the edge weights of the graph, i.e., n-link and t-link potentials \citep{conf/iccv/Boykov2001}.
This is achieved by a convolutional neural network (CNN), which learns features from the images \citep{conf/nips/Krizhevsky2012}.
Here, we do not directly compute these weights solely based on the intensity similarity, as the conventional graph-cuts methods do.
This is because the enhancement patterns in LGE MRI are complex and can vary greatly across different patients, leading to inconsistent intensity patterns.
Also, currently the automatic methods could have a few millimeters under or over segmentation, leading to the estimated endocardial surface being misaligned to the ground truth.
The proposed CNN scheme can exploit both image features and spatial context by means of neighborhood information,
to provide more accurate estimation of the graph weights.
We finally obtain the classification based on the graph-cuts framework, which mitigates the effect of misalignments of the endocardial surface due to automatic LA segmentation errors.

Furthermore, we propose to employ the multi-scale patch (MSP) strategy \citep{journal/mia/Zhuang2016}, and combine it with the CNN for graph potential learning.
This is because distinguishing scars can be challenging solely based on local texture information, particularly in the area where the LA wall is surrounded by other enhanced regions, such as the fibrosis of mitral valve, aortic wall and right atrial (RA) wall.
The MSP method is developed from the image scale space theory, which can handle information at different levels within a limited window and has been widely applied to the tasks of feature extraction, detection and image matching \citep{conf/cg/Lorensen1987}.
The MSP strategy can incorporate both the local fine texture features and the global structural information into the CNN architecture.
We refer to such MSP-based CNN as multi-scale CNN, i.e. MS-CNN.
In addition, the MSPs are extracted with random offsets along the perpendicular direction of the LA endocardial surface, simulating the misalignments between the automatically segmented LA surface and the ground truth.
Therefore, such patches not only can model the multi-scale texture patterns of the images, but also can further improve the robustness of the proposed method against the LA segmentation errors.

\begin{figure*}[thb]\center
 \includegraphics[width=0.7\textwidth]{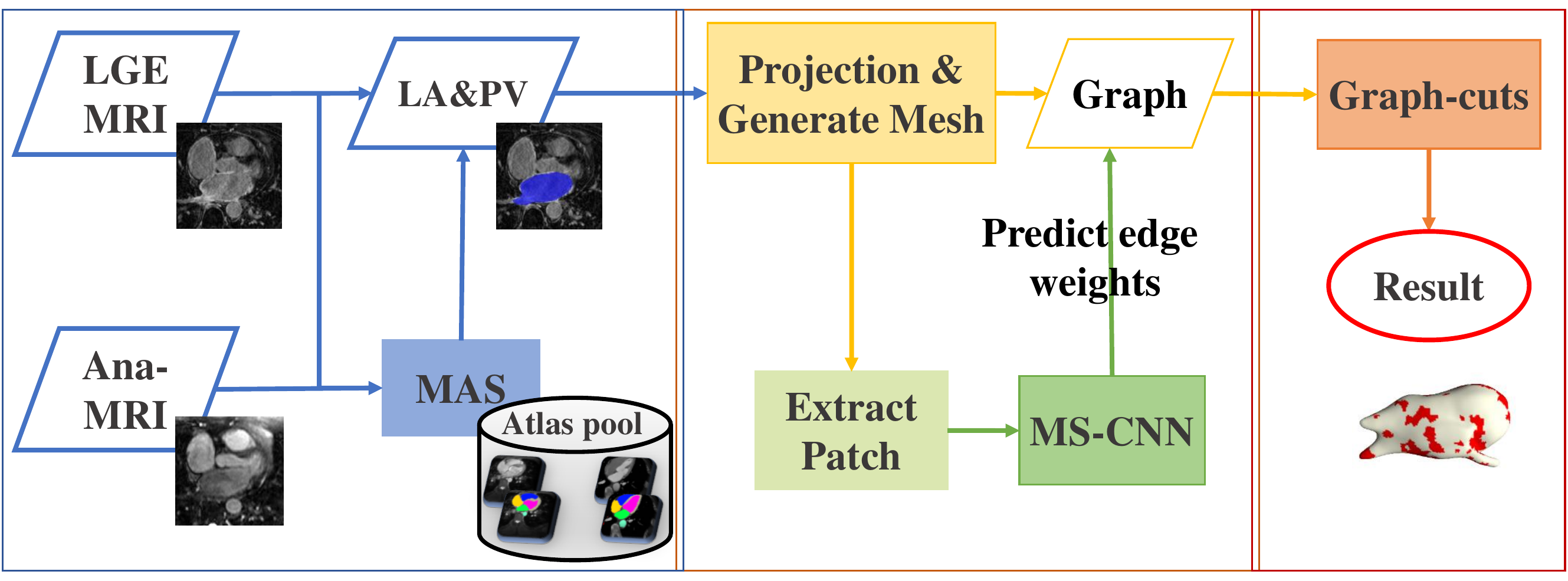}
   \caption{Flowchart of the proposed framework for LA scar quantification and analysis.}
\label{fig:framework}\end{figure*}

The remainder of the paper is organized as follows: the detailed framework of the proposed algorithm is presented in Section \ref{method}. Section \ref{exp} presents the experiments and results. Discussion and conclusion are given in Section \ref{conclusion}.

\section{Method} \label{method}

\zxhreffig{fig:framework} provides an overview of the proposed framework.
First, we use a well-developed multi-atlas whole heart segmentation (MA-WHS) to obtain an initial segmentation of the LA (see Section \ref{method:MAS}).
Then, we project the LA endocardium to generate a surface mesh, where the quantification is performed (see Section \ref{method:projection}).
The labeling of scars is achieved by optimizing a cost function based on the graph-cuts framework (see Section \ref{method:graph-cuts}), whose potentials for edge weights are explicitly learned by the proposed MS-CNN (see Section \ref{method:DNN}).
Note that the graph-cuts based classification is performed on the surface mesh.
This can avoid the challenging segmentation of thin LA wall and also greatly reduce the computational cost.
At the same time, both the texture and anatomical features of the LA myocardium can be adequately extracted by employing the MSP strategy.
Thus, the features of the nodes in the graph are represented by a set of MSPs, and the potentials are learned and predicted by the MS-CNN.

\subsection{Initialization of atrial endocardium and pulmonary veins} \label{method:MAS}
We use MA-WHS, which is based on multi-atlas segmentation (MAS), to obtain the geometrical information of the LA.
This is because the LGE MRI covers the whole heart, and MA-WHS has been well developed and applied in recent years \citep{journal/mia/Zhuang2016,journal/mp/Yang2018}.
MAS algorithm segments an unknown target image by propagating and fusing the labels from multiple annotated atlases using registration.
As the LGE MRI images could have relatively poor image quality, we first apply MA-WHS on the anatomical MRI (Ana-MRI), and then propagate the segmentation using affine registration from the Ana-MRI to LGE MRI.
The Ana-MRI image is normally acquired in the same MRI examination as LGE MRI, using the b-SSFP sequence, which generates higher quality images for atlas-based segmentation.

\begin{figure*}[t]\center\begin{tabular}{@{}c@{}}
 \includegraphics[width=0.98\textwidth]{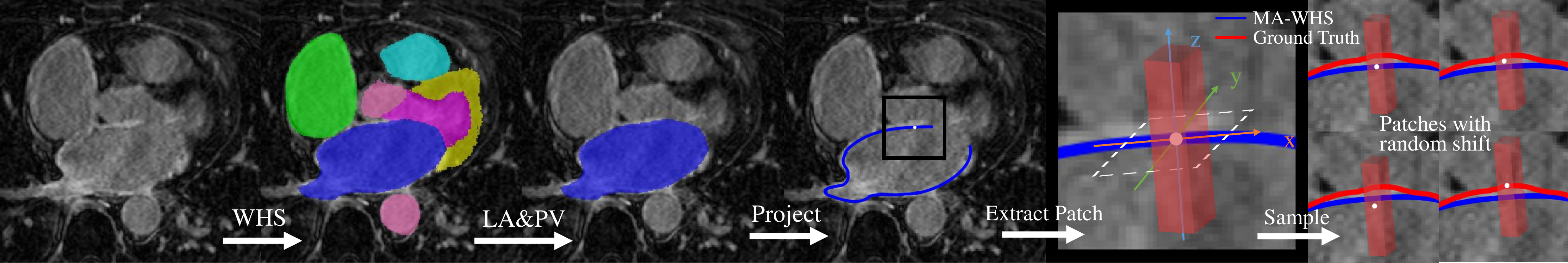}\\
 \makebox[0.98\textwidth][s]{\footnotesize \ (a) \ (b) \ (c) \ (d) \ (e) \ (f)\ } \end{tabular}
   \caption{Pipeline of projection and patch extraction phases. Please refer to the text for more details.}
\label{fig:MAS_patchshift}\end{figure*}

Having finished the WHS for LA and PV delineation,
the marching cubes algorithm \citep{conf/cg/Lorensen1987} is then used to obtain a surface mesh of the LA endocardium which excludes the mitral valve.
Note that the LA segmentation is generally reliable, but still contains errors leading to misalignments between the extracted surface mesh and the ground truth.
For example, the mean Dice score of our MA-WHS for LA is $ 0.898 \pm 0.044 $ (please c.f. Section~\ref{exp_LAseg} for details).
However, the effect of inaccurate LA segmentation can be minimized thanks to the projection strategy and the MS-CNN learning coupled with the randomly shifted MSP sampling strategy.
The reader is referred to \zxhreffig{fig:MAS_patchshift} for illustration and following methodology sections for details.

\subsection{Projection of the atrial endocardium} \label{method:projection}

We project the LA endocardium onto a surface mesh, and then the atrial scars can be classified on a graphical surface.
This is because the clinical demands for scar quantification in AF patients mainly concern the location and extent of scarring areas \citep{journal/tmi/Ravanelli2014}.
\citet{journal/jice/Williams2017} proposed a method to simultaneously represent multiple parameters on a surface model based on the template of an average LA mesh.
By projection, both the errors due to LA wall thickness and misregistration of the WHS can be mitigated.
At the same time, the computational complexity of the algorithm can be reduced dramatically.

The endocardial surface is generated from the volumetric binary segmentation result of the LA cavity using the marching cubes algorithm \citep{conf/cg/Lorensen1987}.
The resolution of the surface mesh is denser than the resolution of the image, which protects the small scars.
The projection from the LA geometry to the surface mesh can preserve the geodesic distances between two nodes.
This equidistant projection is required due to the definition of n-link weights in the proposed graph-cuts framework.
In this formulation, each vertex on the surface, i.e., node of the graph, should include a profile that represents the texture information of the corresponding location in the LGE MRI.
Here, we represent this profile using MSPs, which can incorporate both global structural features and local texture information.

\subsection{Graph formulation for scar segmentation} \label{method:graph-cuts}
Classification and quantification of scars on the LA surface can be formulated as an energy minimization problem solved via graph-cuts.
The weights of the graph come from two parts, i.e., the regional term $E_R$ and the boundary term $E_B$ \citep{conf/iccv/Boykov2001}.
The regional term encodes the intensity distributions of different classes, and the boundary term maintains the continuity between neighbors.

Let $G=\{\mathcal{X},\mathcal{N}\}$ denotes a graph, where $\mathcal{X}=\{x_i\}$ indicates the set of graph nodes, and $\mathcal{N}=\{<x_i,x_j>\}$ is the set of edges.
Here, the weights of edges connecting graph nodes to the terminals are known as t-link weight, and the weights of edges connecting neighboring nodes are referred to as n-link weight.
The two terminals respectively denote the scars and normal myocardium in our problem, analogous to the foreground and background of the general image segmentation task.
Let $l_{x_i}\in\{0,1\}$ be the label assigned to $x_i$, and $l=\{l_{x_i}|{x_i}\in \mathcal{X}\}$ be the label vector that defines a segmentation.
The segmentation energy is defined as follows,

\begin{equation}\begin{array}{l@{\ }l}
 E(l) &= E_R(l) + \lambda E_B(l) \\
 &=\displaystyle\sum_{x_i\in \mathcal{X}}  W_{x_i}^{t-link}(l_{x_i})
 +\lambda \sum_{(x_i, x_j)\in \mathcal{N}}  W_{\{x_i, x_j\}}^{n-link}(l_{x_i}, l_{x_j}).
\end{array}
\end{equation}
where $W_{x_i}^{t-link}$ and $W_{\{x_i, x_j\}}^{n-link}$ are respectively the t-link and n-link weight.

In conventional graph-based segmentation, the regional term is generally obtained by optimizing based on a manual defined initial model.
For example, \citet{conf/iccv/Boykov2001} manually selected a number of seed points to construct such model, referred to as graph cuts method,
and \citet{journal/atg/Rother2004} manually defined a bounding box for interactive segmentation, known as GrabCut approach.
The boundary term in these works was normally defined according to the dissimilarity of intensity and distance between two connected nodes.
\citet{journal/mia/Veni2017} designed a regional term based on a generative image model incorporating both local and global shape priors.
The boundary term was defined for regularizing the smoothness of the estimated surface, i.e., minimizing the squared difference of the offsets between neighboring vertices.
\citet{journal/jcars/Lu2017} estimated a regional term combining three maps, including a probability map, a thresholding map and a local appearance map.
The boundary term they defined was related to the intensity difference and distance of two connected nodes.

In this work, we propose to directly learn and predict the t/n-link potentials for the regional and boundary terms.
This is different from the conventional means, where the profile of a graph node is commonly represented by the intensity of a single pixel or its local texture, which consists of limited information.
Here, we combine the profile representation of graph nodes with the MSP strategy, and learn the potentials using the proposed MS-CNN.
\Zxhreffig{fig:graph} illustrates the flowchart of constructing the graph.

\subsection{Explicit learning of graph potentials using MS-CNN} \label{method:DNN}

\begin{figure*}[t]\center
 \!\!\includegraphics[width=0.70\textwidth]{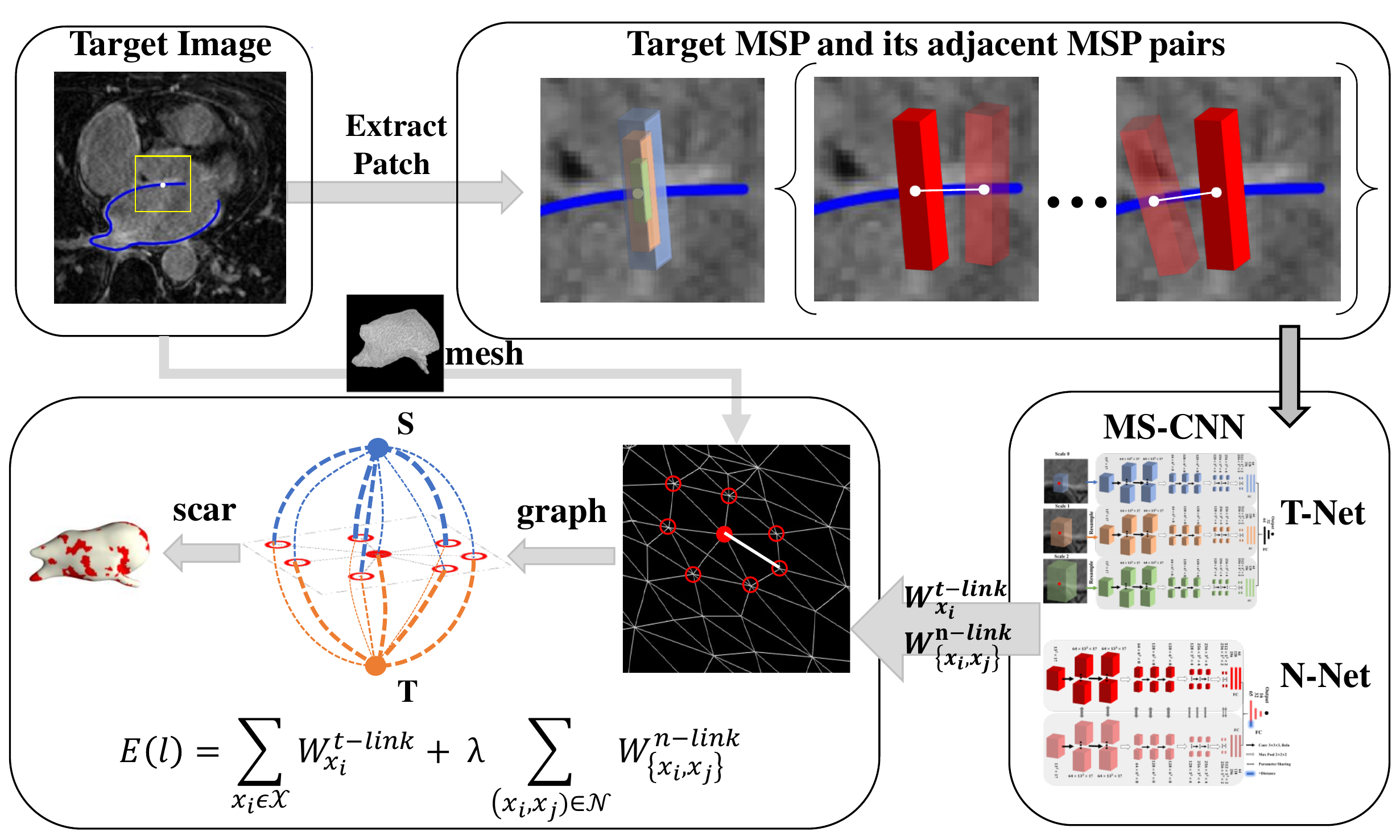}
   \caption{Construction of the graph and the explicit learning of the graph potentials by MS-CNN (MSPs are integrated and represented using red cuboid in this work).}
\label{fig:graph}\end{figure*}

\Zxhreffig{fig:graph} illustrates the computation of graph potentials for the graph-cuts based classification of LA scars.

\subsubsection{Multi-scale patch and patch extraction}
We propose to extract MSPs from LGE MRI to represent the profile of the graph node, and to feed the MS-CNN for training and prediction.
MSP can represent different levels of structural information at a location in an image, with low scale capturing local fine details and high scale providing global structural information of the image \citep{journal/mia/Zhuang2016}.

Each graph node ${x_i}$ has its associated MSPs, denoted as $\mathcal{P}_i=\{p_{x_i}^0, p_{x_i}^1,...,p_{x_i}^{N_s-1}\}$, where $N_s$ indicates the number of scales.
They are extracted from the corresponding volumetric region in the LGE MRI, by back projecting the node to the position in the image.
These patches are elongate-shaped and are defined along the normal direction of the LA endocardial surface, as \zxhreffig{fig:MAS_patchshift} (e) shows,
and their local orientations are maximally aligned to the common world coordinate system of the LGE MRI.
The multi-scale strategy is implemented by adjusting the sample spacing to generate patches with different scales, corresponding to different resolutions of the LGE MRI.
We employ parallel convolutional pathways for multi-scale processing, to feed the different scale information of images to the neural network simultaneously, as \zxhreffig{fig:NET} (a) shows.

\subsubsection{Multi-scale convolutional neural network}
\begin{figure*}[t]\center
    \subfigure[] {\includegraphics[width=0.45\textwidth]{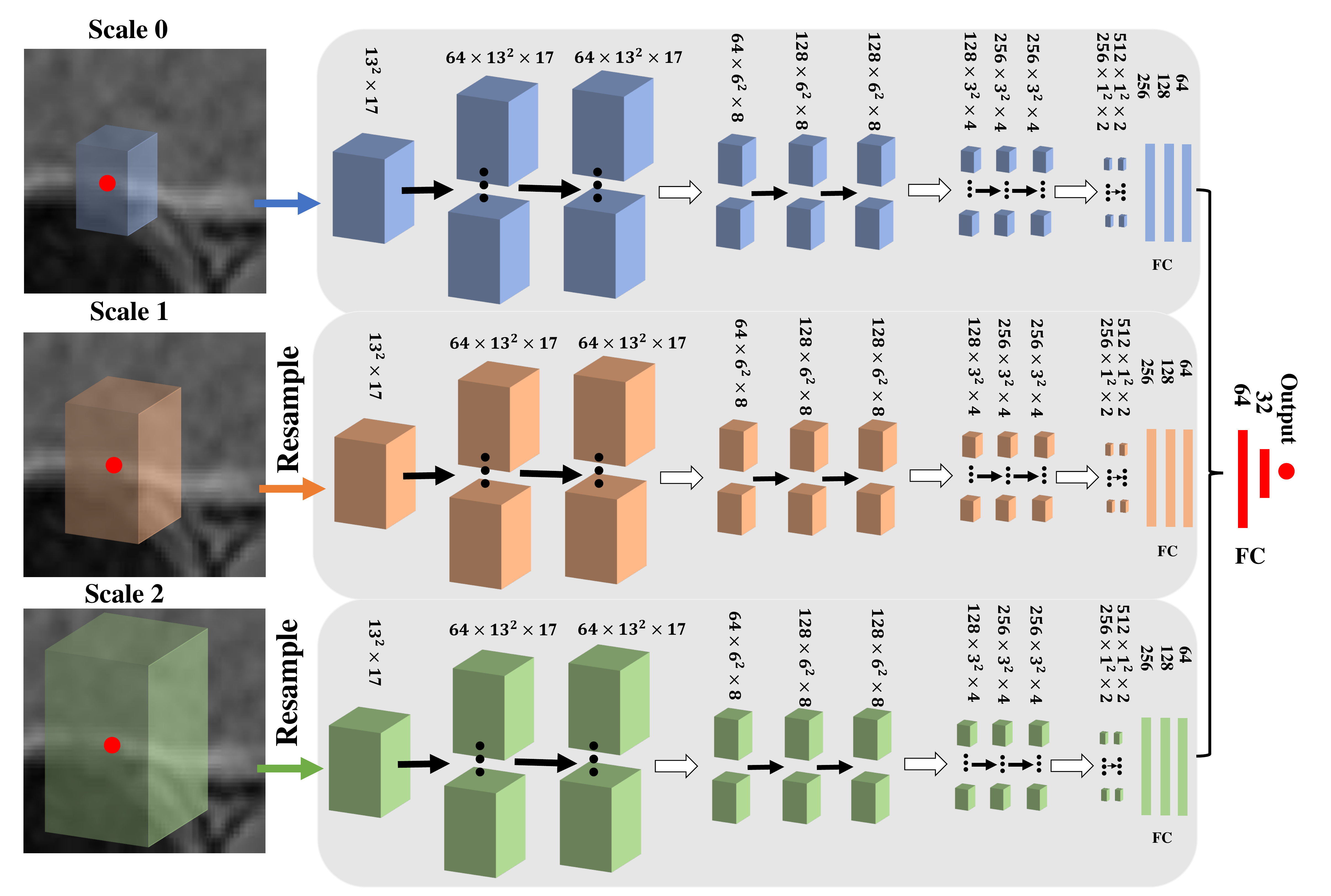}}
    \subfigure[] {\includegraphics[width=0.51\textwidth]{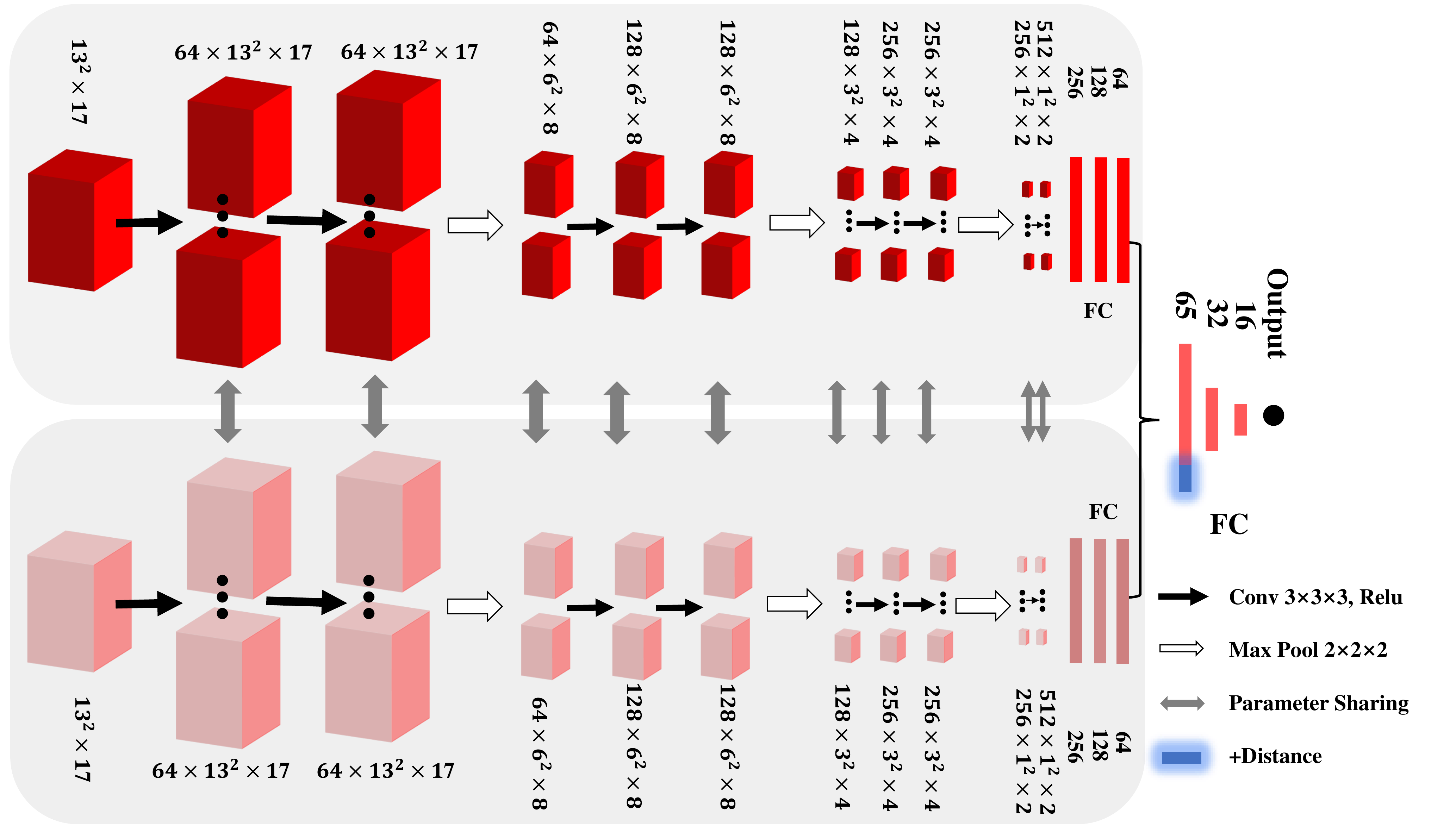}}
   \caption{The hierarchical architecture of the networks:
     (a) $T$-NET;
     (b) $N$-NET.}
\label{fig:NET}\end{figure*}

We have two neural networks, i.e., $T$-NET and $N$-NET.
$T$-NET learns and predicts the t-link potentials, i.e., the probabilities of a node belonging to scars and normal walls respectively, as \zxhreffig{fig:NET} (a) shows.
$N$-NET calculates the n-link potential between two connected nodes, as \zxhreffig{fig:NET} (b) shows.

For training of the t-link potentials, we define a sample for each node of a graph constructed from LGE MRI.
The sample is composed of the MSPs associated to the node $x_i$, and its ground truth label $L_i$, i.e., $L_i$ equals to 1 if it is scar, and 0 otherwise.
As \zxhreffig{fig:NET} (a) shows, the training data of $T$-NET can be represented as $\mathcal{D}^\mathbf{T}=[(\mathcal{P}_1,L_1),...,(\mathcal{P}_N,L_N)]$, i.e., $N$ nodes with corresponding labels.
Thus, the $T$-NET can be parameterized by ${\theta}^\mathbf{T}$ as follows,
\begin{equation}
  \hat{\theta}^\mathbf{T}=\mathop{\arg\min}_{{\theta}^\mathbf{T}} \sum_{i=1}^N (\hat{L}(\mathcal{P}_i;{\theta}^\mathbf{T})-L_i)^2,
\end{equation}
where $\mathcal{P}_i=\{p_{x_i}^0, p_{x_i}^1, p_{x_i}^2\}$, and
$\hat{L}$ is the estimated t-link weight.

For training of the n-link, we define a sample for each pair of two neighboring nodes $\left\{ x_i, x_j \right\}$, consisting of three elements, i.e., (1) the pair of the two sets of MSPs associated with the two nodes, i.e. $\left\{\mathcal{P}_i, \mathcal{P}_j\right\}$, (2) the geodesic distance between them, denoted as $d_{ij}$, and (3) their ground truth label similarity $M_{ij}$.

As \zxhreffig{fig:NET} (b) shows, the training data of $N$-NET can be represented as $\mathcal{D}^\mathbf{N}=[(\mathcal{P}_i, \mathcal{P}_j, d_{ij}, M_{ij})]_{i,j=1}^{i,j=N}$.
The distance $d_{ij}$ is viewed as an additional similarity feature, namely the labeling of two nodes can be more similar if they are closer.
To this end, we design a sub-network, denoted as $\vec{\mathbb{F}}$, to extract high-level and dense features, i.e. $\vec{\mathbb{F}}(\mathcal{P})$.
We then obtain a new feature vector from $\mathcal{P}_i$ and $\mathcal{P}_j$, as follows,
\begin{equation}
  \vec{\mathbb{G}}_{ij}=\vec{\mathbb{F}}(\mathcal{P}_i) \times \vec{\mathbb{F}}(\mathcal{P}_j) + (1-\vec{\mathbb{F}}(\mathcal{P}_i))\times (1-\vec{\mathbb{F}}(\mathcal{P}_j)).
\end{equation}
Each element of $\vec{\mathbb{G}}_{ij}$ can be considered as a similarity metric in the feature space.
Finally, we combine $\vec{\mathbb{G}}_{ij}$ and $d_{ij}$, and feed them to another sub-network for computing the label similarity, i.e., the n-link weight.
Thus, the $N$-NET can be parameterized by ${\theta}^\mathbf{N}$ as follows,
\begin{equation}
  \hat{\theta}^\mathbf{N}=\mathop{\arg\min}_{{\theta}^\mathbf{N}} \sum_{i,j=1}^N (\hat{M}(\vec{\mathbb{G}}_{ij}, d_{ij}; {\theta}^\mathbf{N})-M_{ij})^2,
\end{equation}
where $\hat{M}$ is the estimated n-link weight.

\subsubsection{Training and testing strategy}\label{method:trainingtesting}
In the training phase, we use weighted sampling to mitigate the problem of class imbalance in the training set,
where the number of the nodes belonging to normal myocardium in a subject could be tens or even hundreds times more than that of scars.
In addition, we add a random shift, along with the normal direction, to the center of the MSPs, to mitigate the effects from the inaccurate delineation of the LA boundaries due to over or under segmentation.
This is illustrated in \zxhreffig{fig:MAS_patchshift} (d-f).
This shift should be large enough to overcome potential segmentation errors, while at the same time be small enough to avoid being too distant and cannot capture the texture profile of the LA wall.
We propose to assign this random value in a given range, i.e. $\gamma\in(-\mathcal{R},+\mathcal{R})$, to a node in the training phase, where $\gamma$ is the shift value, $-$ and $+$ represent being inside and outside of the LA blood cavity, respectively.

In the testing phase, one can compute the t-link and n-link potentials of the graph,
and the classification of scars on the LA surface can be achieved by embedding these estimated weights into the graph-cuts framework, i.e.,
\begin{equation}
  W_{x_i}^{t-link}=\hat{L}(\mathcal{P}_i^\gamma;{\theta}^\mathbf{T}),
\end{equation}
and,
\begin{equation}
  W_{\{x_i, x_j\}}^{n-link}=\hat{M}(\mathcal{P}_i^\gamma,\mathcal{P}_j^\gamma,d_{ij};{\theta}^\mathbf{N})=\hat{M}(\vec{\mathbb{G}}_{ij},d_{ij};{\theta}^\mathbf{N}).
\end{equation}
Note that the two normalized t-link weights of a node, indicating potentials to the foreground and background respectively, can also be viewed as the probabilities of this node belonging to scars and normal tissues.

\section{Experiments and results} \label{exp}

\subsection{Data acquisition and experimental setup}
We collected fifty-eight post-ablation LGE MRI data from patients with longstanding persistent AF for experiments.
Transverse navigator-gated 3D LGE MRI was performed on a 1.5T Siemens Magnetom Avanto scanner (Siemens Medical Systems, Erlangen, Germany), which used an inversion prepared segmented gradient echo sequence (TE/TR 2.2 ms/5.2 ms) 15 minutes after gadolinium administration.
The LGE MRI data were acquired at resolution of (1.4-1.5) $\!\times\!$ (1.4-1.5) $\!\times\!$ 4 mm, and reconstructed to (0.7-0.75)$\!\times\!$ (0.7-0.75) $\!\times\!$ 2 mm.
For each patient, prior to contrast agent administration, coronal navigator-gated 3D b-SSFP (TE/TR 1 ms/2.3 ms) data were scanned, with acquisition resolution of (1.6-1.8) $\!\times\!$ (1.6-1.8) $\!\times\!$ 3.2 mm, and reconstructed to (0.8-0.9) $\!\times\!$ (0.8-0.9) $\!\times\!$ 1.6 mm.
Both LGE MRI and b-SSFP data were acquired during free breathing with respiratory motion control \citep{journal/mrm/Keegan2014}.

The available data were randomly divided into two sets, one for training (31 images) and the other for testing (27 images).
$T$-NET was trained using stochastic gradient descent optimizer, with following hyper-parameters: momentum = 0.9, batch size=50, weight decay=$10^{-4}$, number of epochs=15.
The learning rate was initially set to 0.01, and had a stepped decay rate of 0.8 every 1000 iterations.
A similar configuration was designed for $N$-NET.

We first evaluated the accuracy of automatic segmentation of LA in Section~\ref{exp_LAseg}.
Then, we performed four parameter studies to verify the effects of the parameters and explore their optimal values.
In Section~\ref{exp_patchsize}, we investigated the influence of different patch sizes to the proposed framework using the single-scale CNN, and then compared the results with that of MS-CNN.
In Section~\ref{exp_lambda}, we studied the proposed method with different values of the balancing parameter $\lambda$.
Section~\ref{exp_randomshift} and Section~\ref{exp_scale} present the studies of random shift and multi-scale learning, respectively.
The optimal parameters concluded from these studies were used for the proposed method, in comparisons with other methods, in Section~\ref{exp_comparison}.
Finally, Section~\ref{result_performance} reports the performance of the proposed method and results of the inter-observer study.

\subsection{Gold standard and evaluation} \label{evaluation}

All the LGE MRIs were manually segmented by an experienced cardiologist specialized in cardiac MRI, to label the enhanced atrial scarring regions, which are considered as ground truth in this work.
To assess the scar classification results, we generated the ground truth reference by projecting the manually segmented scars onto the LA surface.
With regard to different initializations, i.e., the manual (abbreviated as $LA_M$) and automatic (abbreviated as $LA_{auto}$) delineation of LA, two different ground truths, respectively referred to as $GT_M$ and $GT_{auto}$, were generated for evaluation.
In the comparison studies in Section \ref{exp_comparison}, the fully automatic methods were evaluated using $GT_{auto}$, while the semi-automatic algorithms based on $LA_M$ were evaluated using $GT_M$.

For evaluation, we computed the statistical measures, Dice score of scars, referred to as Dice (scar), and the generalized Dice score, denoted as $G$Dice.
The statistical measures include accuracy, sensitivity and specificity.
$G$Dice is a weighted Dice score by evaluating the segmentation of all labels \citep{journal/tmi/CrumCH2006,journal/jhe/Zhuang2013}, and is formulated as follows,
\begin{equation}\begin{array}{l@{\ }l}
 G$Dice$ &= \frac {2\sum_{k=0}^{N_{k}-1}\left| {S}_{k}^{auto} \cap {S}_{k}^{manual}\right|} {\sum_{k=0}^{N_{k}-1}(\left| S_{k}^{auto}\right|) + (\left|S_{k}^{manual}\right|)},
\end{array}
\end{equation}
where $S_{k}^{auto}$ and $S_{k}^{manual}$  indicate the segmentation results of label $k$ from the automatic method and manual delineation, respectively, and $N_{k}$ is the number of labels.
All the metrics are computed on the projected LA surface.

\subsection{Automatic segmentation of LA and correlation analysis} \label{exp_LAseg}
To obtain an initialization of LA for scar segmentation, we developed the MA-WHS method using 30 b-SSFP MRI atlases.
The 30 high resolution atlases were constructed from the Left Atrial Segmentation Challenge (STACOM 2013) \citep{journal/tmi/tobon2015}.
The manual delineation of LA was regarded as the gold standard for this experiment.
The MA-WHS results of Ana-MRI were mapped to LGE MRI from the same subject, and then generated the initial LA labels.
The average Dice score of this LA segmentation to the manual delineation was $ 0.898 \pm 0.044 $.

\begin{figure}[t]\center
 \includegraphics[width=0.46\textwidth]{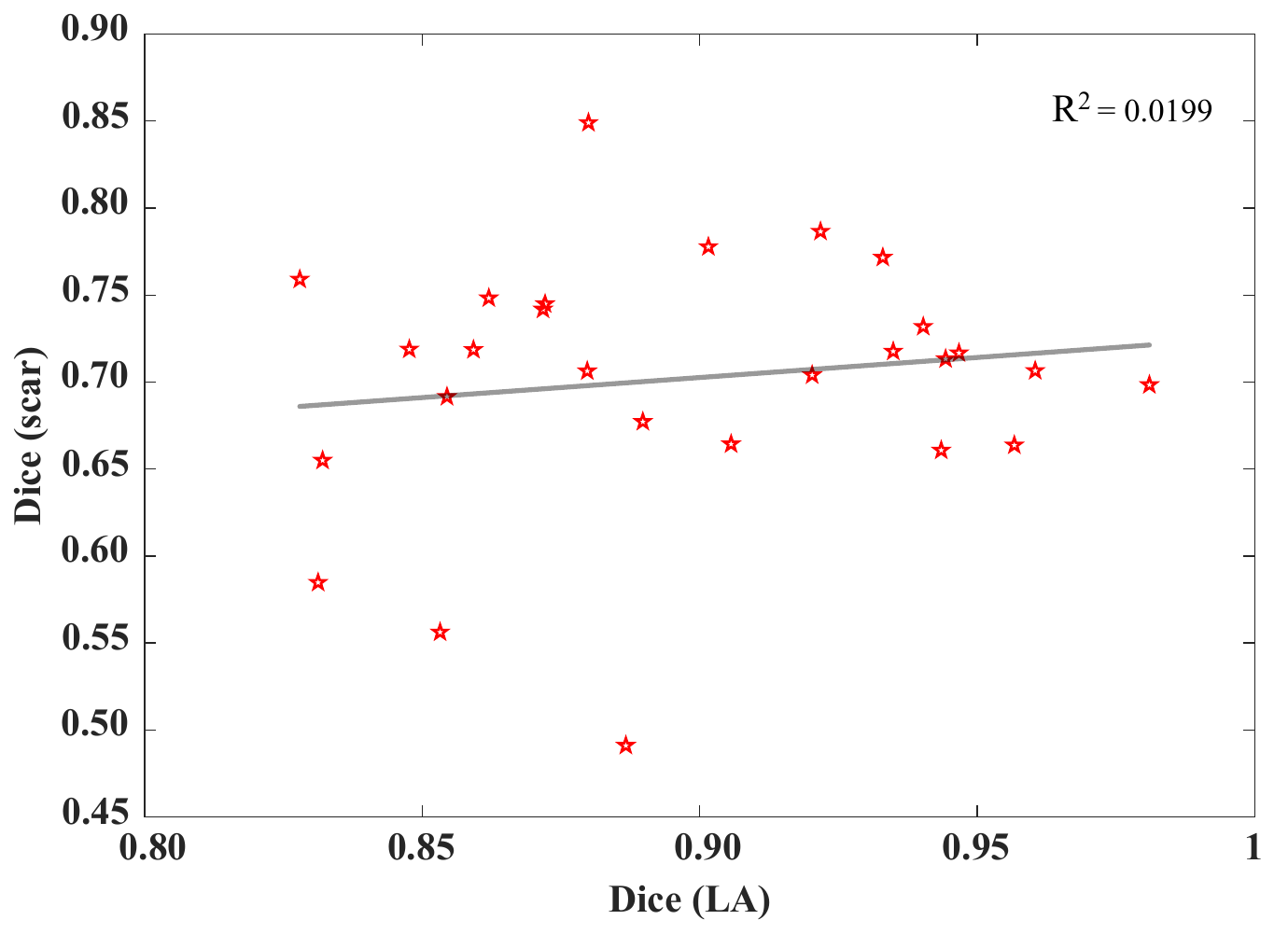}
   \caption{Scatter point plot for analyzing the correlation between the LA segmentation performance and scar quantification accuracy, both indicated by Dice scores.
   The Pearson coefficient and Spearman's rank coefficient are respectively 0.1412 and 0.0110.}
\label{fig:exp:correlation}\end{figure}

To analyze the relation between the LA segmentation error and the scar quantification accuracy by the proposed method,
we plotted these two values for each of the 27 test subjects as two dimension scatter points in \zxhreffig{fig:exp:correlation}.
One can see that the plot shows little direct relationship between them.
We further performed linear regression, Pearson correlation and Spearman's rank correlation.
The $R^2$ and Pearson coefficient were respectively 0.0199 and 0.1412, indicating low linear correlation between Dice (scar) and Dice (LA);
and the rank correlation coefficient was 0.0110, meaning hardly monotonic relationship between them either.
To conclude, the result illustrates the low correlation between the scar quantification accuracy and the LA segmentation accuracy by the proposed method.

\subsection{Parameter Studies}

\begin{figure*}[t]\center
    \subfigure[] {\includegraphics[width=0.46\textwidth]{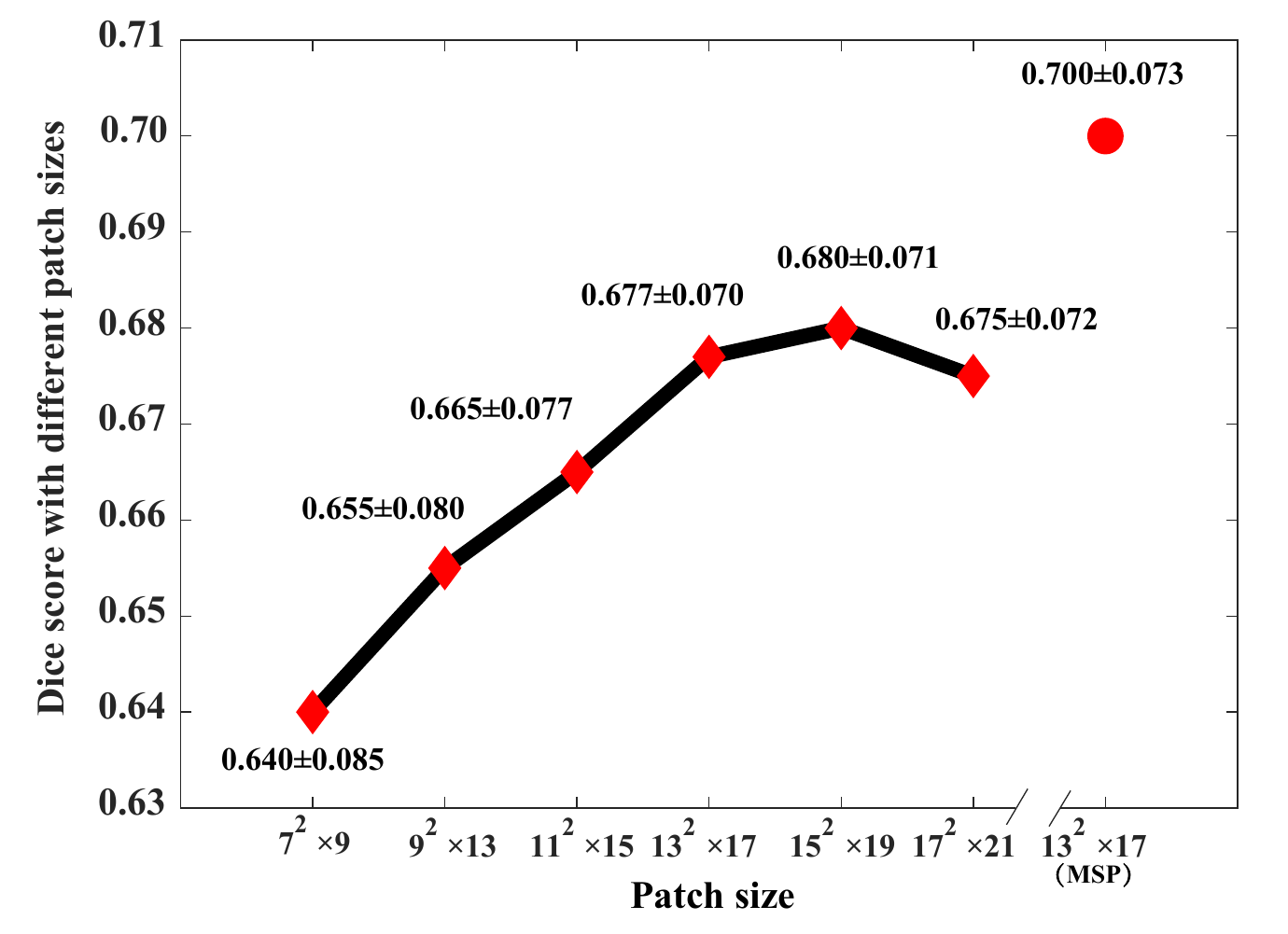}}
    \subfigure[] {\includegraphics[width=0.46\textwidth]{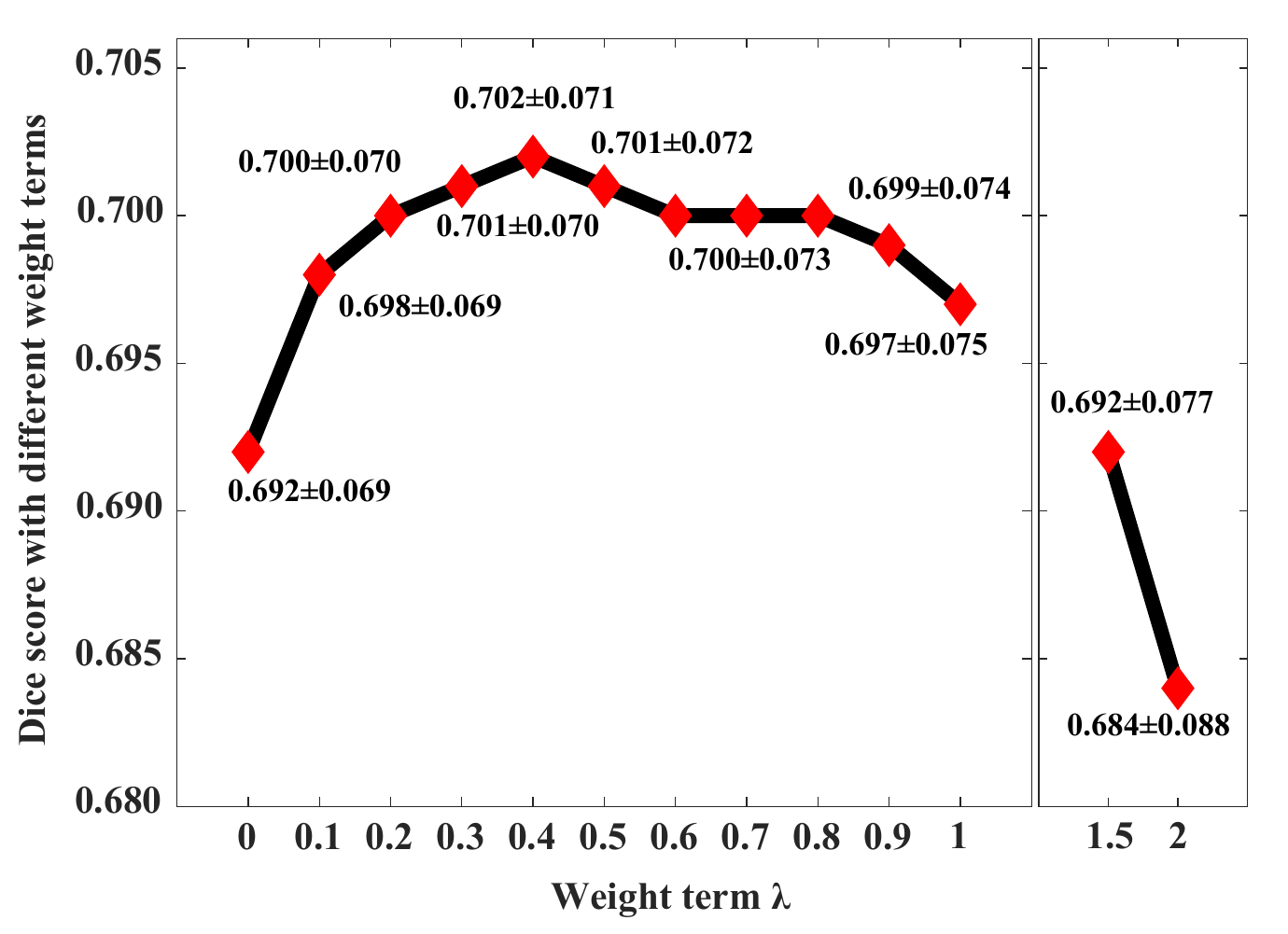}}\\
    \subfigure[] {\includegraphics[width=0.46\textwidth]{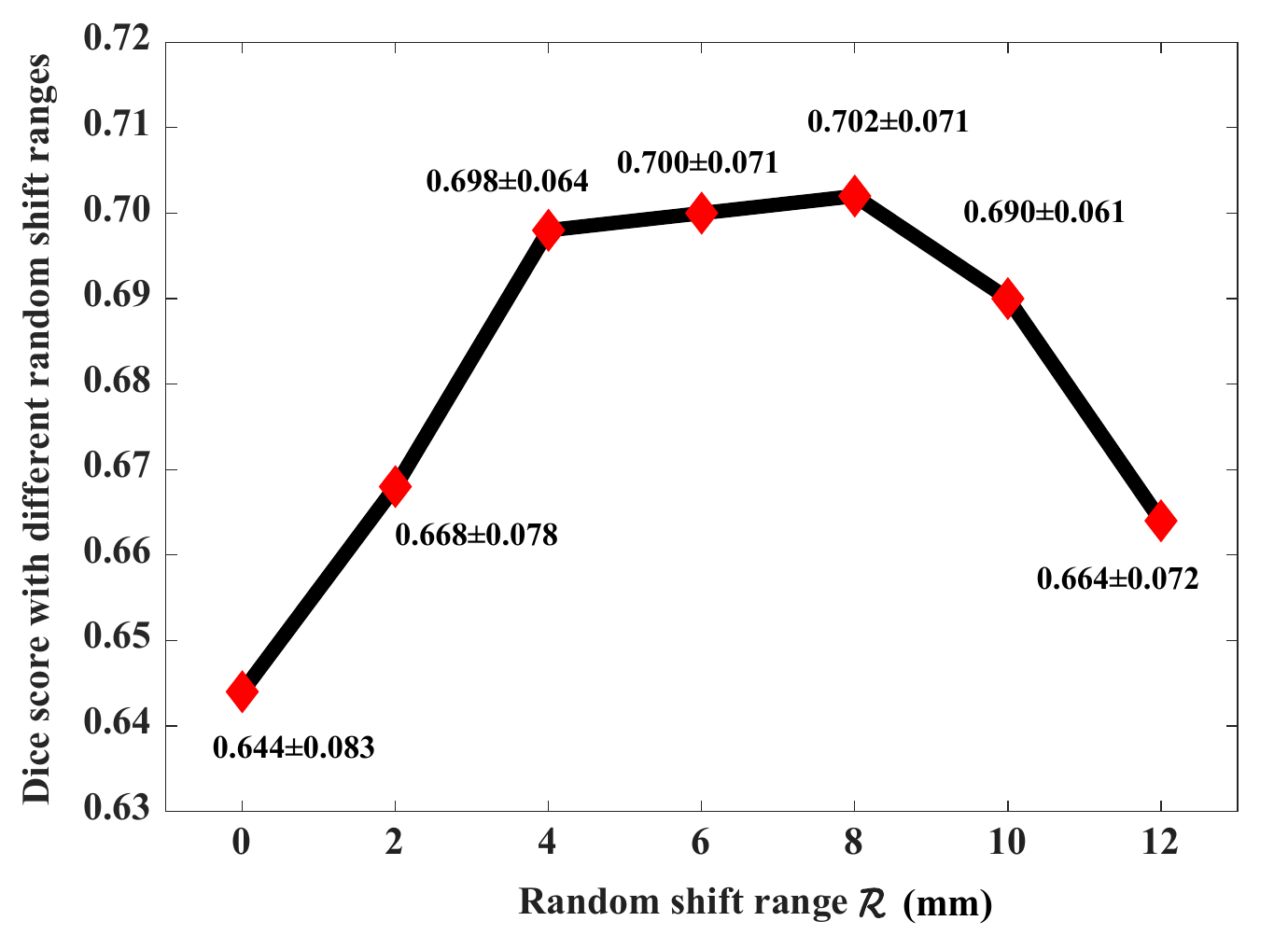}}
    \subfigure[] {\includegraphics[width=0.46\textwidth]{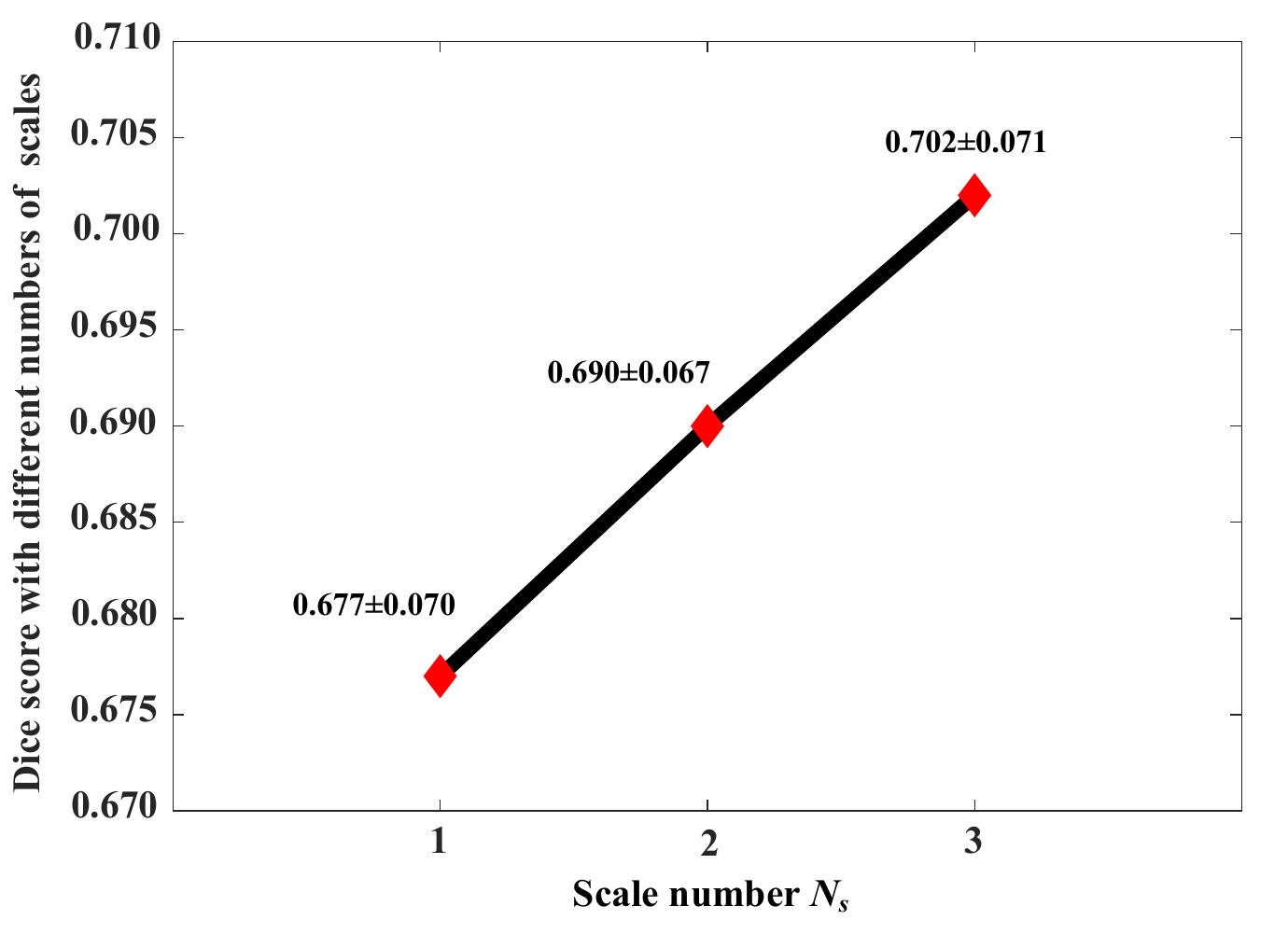}}
   \caption{Dice scores of the proposed method with different parameterizations:
     (a) performance against different patch sizes ($\lambda$=0.6);
     (b) performance against different values of the balancing parameter $\lambda$ to weight the t-link and n-link terms in the graph-cuts framework;
     (c) performance against different random shift ranges $\mathcal{R}$;
     (d) performance against different numbers of scales $N_s$.}
\label{fig:exp:parameter}\end{figure*}

\subsubsection{Study of patch sizes} \label{exp_patchsize}
We used one-scale MSP, namely only the original image (scale 0) was used and the CNN was a single-scale network, for studying the proposed method with different sizes of patches.
The patch sizes ranged from 7 $\!\times\!$ 7 $\!\times\!$ 11 to 17 $\!\times\!$ 17 $\!\times\!$ 21 voxel, where the voxel size is 1 $\!\times\!$ 1 $\!\times\!$ 1 mm.
Then, we implemented the three-scale MSP and CNN with patch size 13 $\!\times\!$ 13 $\!\times\!$ 17 voxel, for comparisons with the single-scale CNNs.
The balancing parameter $\lambda$ in this study was set to 0.6, and the random shift range $\mathcal{R}$ was set to half of the patch length.

\zxhreffig{fig:exp:parameter} (a) shows that the average Dice score increases dramatically at first with respect to the increased sizes of patches, then starts to converge after the patch size reaching 13 $\!\times\!$ 13 $\!\times\!$ 17 voxel.
This is reasonable, as the larger size is used, the richer intensity profile is included for feature training and detection.
However, the increase of patch size generally requires more complex networks, either more kernels or more convolutional layers, which increases computation load and memory requirements.
This also rationalizes our proposal to use MSP and MS-CNN.
As \zxhreffig{fig:exp:parameter} (a) presents, our MS-CNN drastically increases the accuracy of the classification results, thanks to the usage of the MSP strategy which incorporates both local and global information of the images.
In the following experiments, we adopted this three-scale setting (except for Section \ref{exp_scale}) and patch size of 13 $\!\times\!$ 13 $\!\times\!$ 17 voxel.

\subsubsection{Study of balancing parameter $\lambda$} \label{exp_lambda}
In this study, we compared the results of the proposed scheme using different values for the balancing parameter $\lambda$, $\lambda \in {[0, \infty)}$, to demonstrate the effect of graph-cuts.
Here, we set the values ranging from 0 to 2.
The patch strategy was as follows, number of scales was three, patch size was 13 $\!\times\!$ 13 $\!\times\!$ 17 voxel, and the random shift range $\mathcal{R}$ was set to a maximum of 8 mm.

\zxhreffig{fig:exp:parameter} (b) presents the results.
One can see that the best performance in terms of Dice score is obtained when $\lambda$ is set to 0.4.
This indicates that the inter-node relation (n-link) is important, and the weighting between the t-link and n-link terms should be balanced to achieve optimal performance.
In the following experiments, $\lambda$ was set to 0.4 for the proposed method.

\subsubsection{Study of random shift range} \label{exp_randomshift}
To demonstrate the effect of random shift, we compared the performance of the proposed method with different random shift ranges $\mathcal{R}$, for $\gamma\in(-\mathcal{R},+\mathcal{R})$.
Here, we set $\mathcal{R}$ ranging from 0 to 12 mm.
The patch size was 13 $\!\times\!$ 13 $\!\times\!$ 17 voxel, and $\lambda$ was set to 0.4.

\zxhreffig{fig:exp:parameter} (c) provides the results of this study.
The best Dice score is obtained when $\mathcal{R}$ is set to 8 mm, i.e., half of the patch length in the long-axis direction,
 and the performance of the proposed method deteriorates drastically when the shift range becomes larger than 8 mm.
This is rational, because the shift range should cover all the potential misalignments of the constructed surface to the ground truth.
When the random shift range $\mathcal{R}$ is greater than 8 mm, the patch may not cover the regions which include the important features for training and classification.

\subsubsection{Study of scales} \label{exp_scale}
To study the effect of multi-scale learning, we compared the results using different numbers of scales, i.e. $N_s=\left\{1,2,3\right\}$. 
The patch size of MSP was set to 13 $\!\times\!$ 13 $\!\times\!$ 17 voxel, $\lambda$ was set to 0.4, and the random shift range $\mathcal{R}$ was set to a maximum of 8 mm.

\zxhreffig{fig:exp:parameter} (d) presents the mean Dice scores of the method.
This study demonstrates that the effectiveness of the multi-scale learning.
It indicates that the more scales we used the better accuracy we obtained.
It should be noted that when we tried to use more scales, the training session failed, due to the limited computation capacity of our computer.

\subsection{Comparison with other methods} \label{exp_comparison}
In this study, we implemented eight segmentation approaches, including the proposed method, for comparisons.
Here, $LA_M$ indicates the methods adopt the manual segmentation of LA for initialization,
and $LA_{auto}$ denotes the methods employ the automatic segmentation from the MA-WHS approach described in Section \ref{method:MAS}.

\begin{table*} [t] \center
    \caption{
    Summary of the quantitative evaluation results. $G$Dice denotes the generalized Dice score.
    Here, the asterisk ($^*$) in column Dice (scar) indicates the methods obtained statistically poorer ($p<0.01$) results
	compared to the proposed $LA_{auto} + LearnGC$. The $p$ value of the Dice (scar) between $LA_M + MS$-$CNN$$^0$ and $LA_{auto} + MS$-$CNN$$^0$ is 0.225.
     }
\label{tb:table:comp_post}
{\small
\begin{tabular}{ l| l *{4}{@{\ \,} l }}\hline
Method       & \quad Accuracy & \qquad Sensitivity & \qquad Specificity & \qquad Dice (scar)  & \qquad $G$Dice \\
\hline
$LA_M + 2SD$                &$ 0.809 \pm 0.074 $&  \quad $ 0.168 \pm 0.067 $&  \quad \bm{$ 0.994 \pm 0.005 $}&  \quad $ 0.275 \pm 0.091^* $&  \quad $ 0.758 \pm 0.098 $  \\
$LA_M + Otsu$               &$ 0.763 \pm 0.188 $&  \quad $ 0.346 \pm 0.214 $&  \quad $ 0.880 \pm 0.289 $&  \quad $ 0.396 \pm 0.090^* $&  \quad $ 0.726 \pm 0.207 $  \\
$LA_M + MGMM$               &$ 0.708 \pm 0.160 $&  \quad $ 0.781 \pm 0.127 $&  \quad $ 0.690 \pm 0.236 $&  \quad $ 0.545 \pm 0.101^* $&  \quad $ 0.716 \pm 0.190 $  \\
$LA_M + MGMM + GC$          &$ 0.716 \pm 0.162 $&  \quad \bm{$ 0.799 \pm 0.124 $}&  \quad $ 0.694 \pm 0.240 $&  \quad $ 0.562 \pm 0.102^* $&  \quad $ 0.721 \pm 0.192 $  \\
$LA_M + MS$-$CNN^0$         &$ 0.798 \pm 0.051 $&  \quad $ 0.775 \pm 0.099 $&  \quad $ 0.805 \pm 0.078 $&  \quad $ 0.615 \pm 0.083^* $&  \quad $ 0.811 \pm 0.047 $  \\ 
$LA_{auto} + MS$-$CNN^0$    &$ 0.806 \pm 0.052 $&  \quad $ 0.743 \pm 0.126 $&  \quad $ 0.824 \pm 0.088 $&  \quad $ 0.631 \pm 0.080^* $&  \quad $ 0.814 \pm 0.047 $  \\ 
$LA_{auto} + MS$-$CNN$      &$ 0.846 \pm 0.032 $&  \quad $ 0.786 \pm 0.118 $&  \quad $ 0.886 \pm 0.057 $&  \quad $ 0.692 \pm 0.069^* $&  \quad $ 0.851 \pm 0.030 $  \\
$LA_{auto} + LearnGC$       &\bm{$ 0.856 \pm 0.033 $}&  \quad $ 0.773 \pm 0.132 $&  \quad $ 0.883 \pm 0.058 $&  \quad \bm{$ 0.702 \pm 0.071 $}&  \quad \bm{$ 0.859 \pm 0.031 $}  \\
\hline
\end{tabular} }\\
\end{table*}

\begin{enumerate}
 \item[(1)]$LA_M + 2SD$:
   This is one of the most widespread thresholding methods to detect atrial scars.
   It calculates a specific number of standard deviation (SD) above a reference value.
   The reference value is generally set to the mean intensity from the blood pool or LA wall.
   It is however generally patient-specific and slice-specific, and different numbers of SD have been used \citep{journal/jcmr/Karim2013}.
   In our study, we obtained the optimal performance by setting the threshold value to 2 SD above the mean intensity of LA walls.
   Here, we constructed the LA wall from a manual segmentation of the LA with a morphological dilation,
   which was also used for the following experiments when the LA wall was needed from $LA_M$.
 \item[(2)]$LA_M + Otsu$: This method uses the Otsu algorithm \citep{journal/tsmc/Otsu1979} for automatic thresholding of the scarring tissues from the LA wall obtained from $LA_M$.

 \item[(3)]$LA_M + MGMM$: This method adopts the multi-component Gaussian mixture model (MGMM) for scar segmentation from the LA wall \citep{journal/tbe/Liu2017}.
     MGMM can deal with the intensity heterogeneity of myocardium caused by the infarcts, and has been proven to be effective in myocardium segmentation.

 \item[(4)]$LA_M + MGMM + GC$: This method further regularizes the spatial continuity using the graph-cuts framework, based on the result of MGMM.
     Here, we defined the boundary weight using the intensity difference between neighboring points,
     and the regional weight was computed from the posterior probability map of scars generated from MGMM.

 \item[(5)]$LA_M + MS$-$CNN^0$: This learning based method only uses the two t-link weights estimated from $T$-NET to classify scars.
    The two weights, i.e. respectively linked to the foreground scar and background normal tissue, are normalized and considered as the posterior probability of the two labels.
    Here, both training data and test data were initialized using manually segmented LA, so the random shift in the training phase was set to zero, i.e. $\gamma$=0.

 \item[(6)]$LA_{auto} + MS$-$CNN^0$: This method uses the estimated t-link weights from $T$-NET, similar to $LA_M + MS$-$CNN^0$, to classify scars. However, the LA here was automatically segmented using MA-WHS.
     For comparisons with $LA_M + MS$-$CNN^0$, here we also set the random shift to zero ($\gamma$=0).

 \item[(7)]$LA_{auto} + MS$-$CNN$: Similarly, this method uses the estimated t-link weights from $T$-NET to classify scars, and the LA was automatically segmented using MA-WHS.
     However, in the training phase we set the random shift accordingly based on the parameter study in Section~\ref{exp_randomshift}.

 \item[(8)]$LA_{auto} + LearnGC$: This is the proposed method in which the LA was initialized by MA-WHS and the weights of the graph were learned and predicted using MS-CNN.
    Here, the balancing parameter $\lambda$ was set to 0.4. Noted that when $\lambda=0$, $LA_{auto} + LearnGC$ becomes $LA_{auto} + MS$-$CNN$.

\end{enumerate}

\begin{figure}[t]\center
 \includegraphics[width=0.48\textwidth]{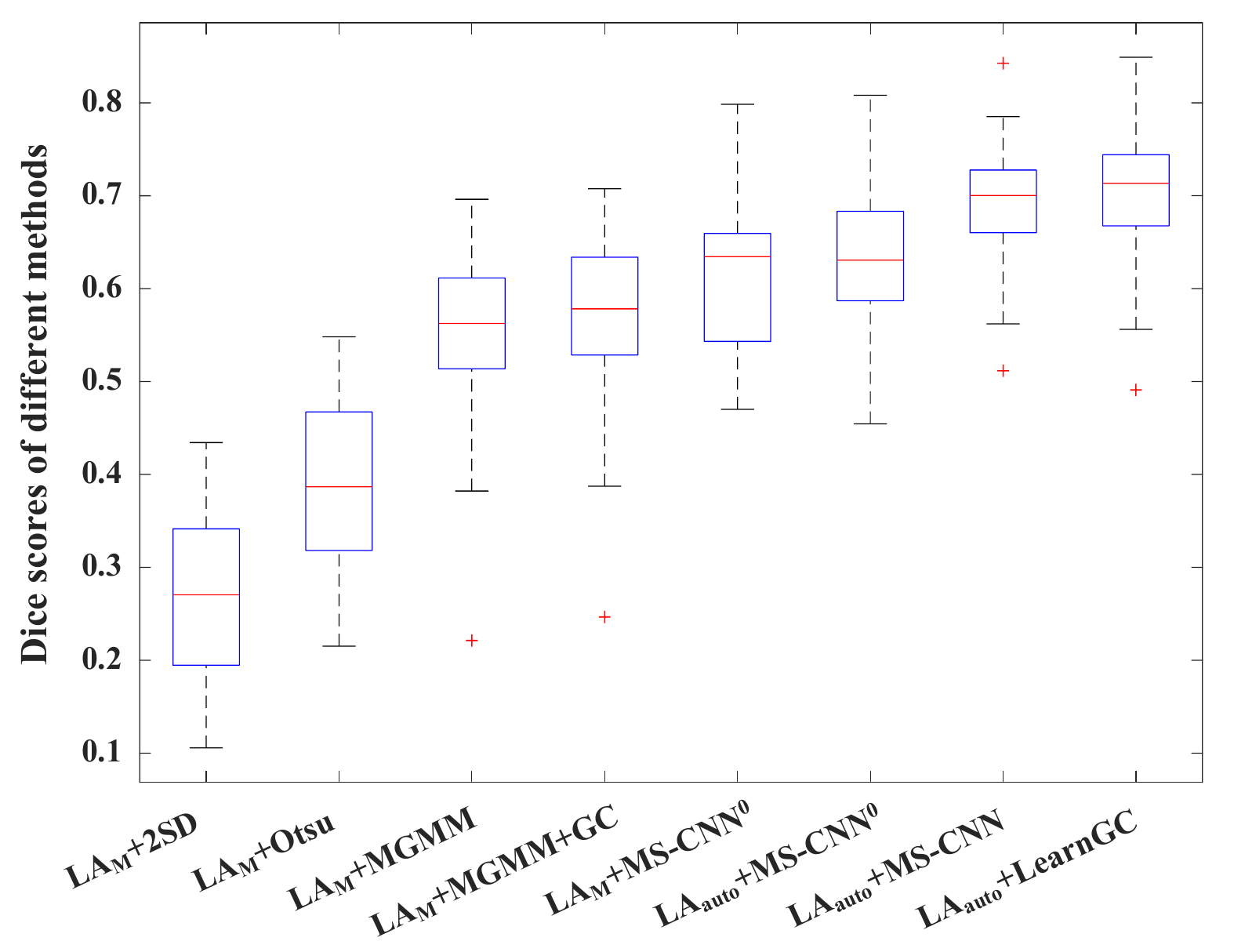}
   \caption{Boxplots of the Dice scores of scars by the eight methods.}
\label{fig:exp:comp}\end{figure}

\begin{figure}[t]\center
  \includegraphics[width=0.5\textwidth]{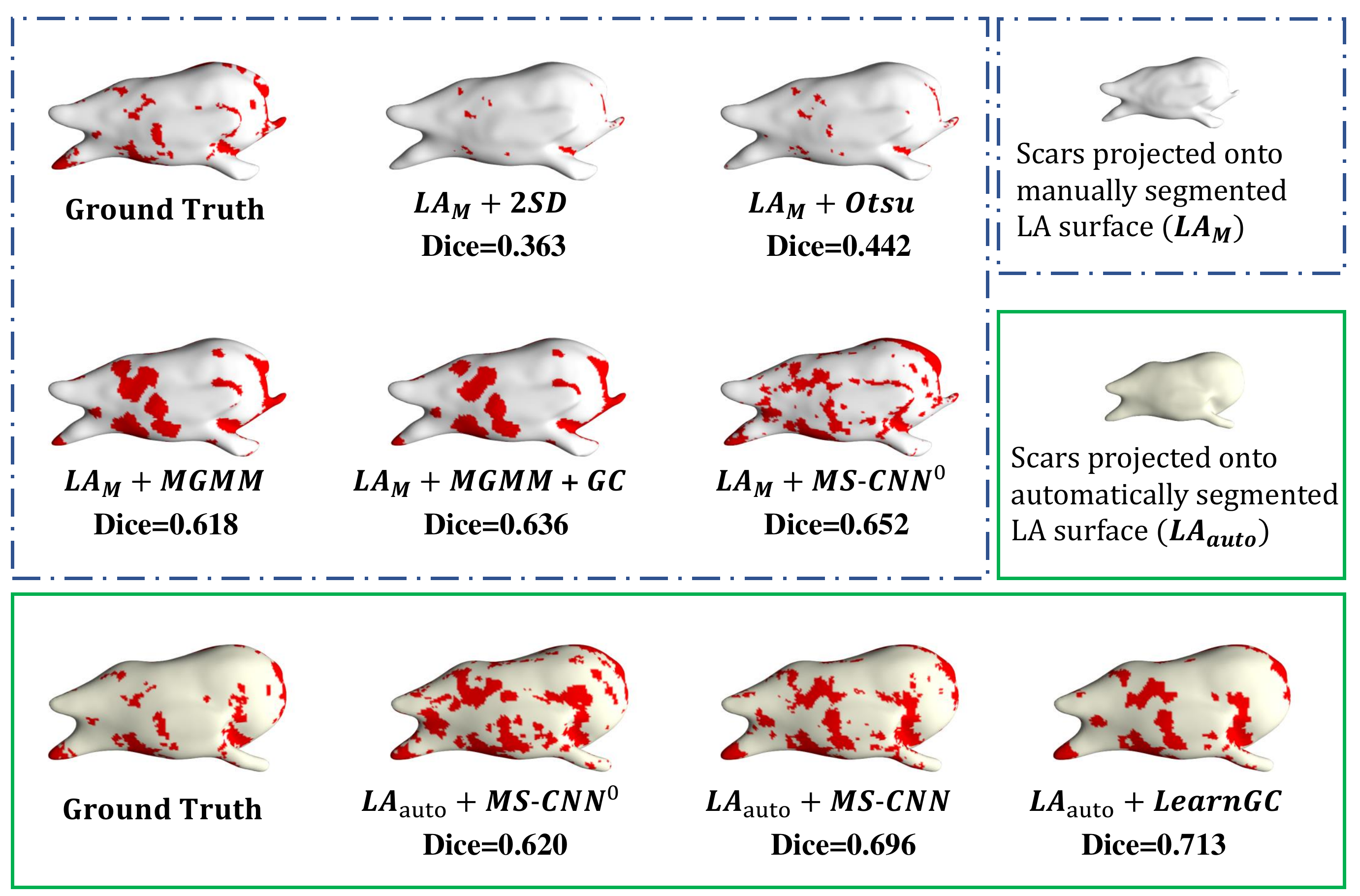}
   \caption{3D visualization of the LA scar classification results using the eight methods.
   This is the median case selected from the test set in terms of Dice score of scars by the proposed method.
   The scarring areas are red-colored on the LA surface mesh, which can be constructed either from $LA_M$ (LA surface in white) or from $LA_{auto}$ (LA surface in light yellow).}
\label{fig:result:3dvisual}\end{figure}

\Zxhreftb{tb:table:comp_post} presents all the quantitative results of the eight methods,
and \zxhreffig{fig:exp:comp} provides their boxplots of Dice scores of scars.
The proposed learning graph-cuts method, i.e. $LA_{auto} + LearnGC$, obtained evidently better scar segmentation (Dice of scars) than the conventional methods based on $LA_M$.
It also performed statistically better than all the other seven methods in terms of Dice scores of scars ($p<0.01$).
Note that $LA_{auto} + MS$-$CNN$ has a slightly better Dice (scar) than $LA_M + MS$-$CNN$ but without statistical significance (p=0.255), even though the former is based on automatic segmentation of LA and the latter uses manual segmentations.
When combined with the random shift strategy, $LA_{auto}+LearnGC$ and $LA_{auto} + MS$-$CNN$ obtained evidently and statistical better Dice (scar) than the other methods ($p<0.01$).
For them, $LA_{auto}+LearnGC$ is generally better, but the gain is marginal, due to the fact that the graph-cuts is considered as a built-in smoothness constraint to generate less patchy results.
In this study, $LA_{auto}+LearnGC$ did not obtain the best figures in \emph{sensitivity} or \emph{specificity} metrics.
Sensitivity measures the proportion of actual scarring regions that are correctly identified, and specificity measures the proportion of actual normal wall regions that are correctly segmented.
One can see the misleading of these two metrics in evaluating the performance of a method from \zxhreftb{tb:table:comp_post}, where $LA_M + 2SD$ and $LA_M + MGMM + GC$ achieved the best specificity or sensitivity, though their performance was actually poor in our visual assessment.

In addition, we chose a representative case, the median in terms of Dice (scar) from the test set by the proposed $LA_{auto}+LearnGC$.
\Zxhreffig{fig:result:3dvisual} visualizes the 3D results by the eight methods.
One can observe that the 3D visualization agrees well with the quantitative analysis result using Dice (scar).
Though the manually segmented scars in $LA_M$ and $LA_{auto}$ are projected onto two different reference surfaces, $GT_M$ and $GT_{auto}$ visually appear similar when we compare the location and extent of scars.
Both the two threshold algorithms, 2SD and Otsu, tended to under estimate (segment) the scars, though Otsu generally performed better.
The results of $LA_M + MGMM$ and $LA_M + MGMM+ GC $ were acceptable, but the accuracy and automation needed improving.
The leaning-based methods, from {$LA_M + MS$-$CNN^0$}, {$LA_{auto} + MS$-$CNN^0$} and {$LA_{auto} + MS$-$CNN$}, to {$LA_{auto} + LearnGC$}, improved the performance when the new methodologies were introduced.
Particularly, {$LA_{auto} + LearnGC$} further reduced the noise and patchy segmentation results, and it obtained full automation and best Dice score of scar  quantification.

\subsection{Performance of the proposed method and inter-observer study} \label{result_performance}
This study analyzes the performance of the proposed method in detail.
To provide a reference for the quantitative evaluation metrics, we conducted a study of inter-observation variation from two manual delineations.
We randomly selected ten cases from the available data, and asked two experts to manually label the scars separately.
For each case, the two labelling results of scars were projected onto the $LA_M$ surface.
The Dice (scar), generalized Dice, and accuracy of inter-observer variation were respectively $ 0.695 \pm 0.049 $, $0.868\pm0.027$ and $0.867\pm0.026$.

\Zxhreftb{tb:table:comp_post} summarizes the quantitative evaluation results of the proposed method, i.e. $LA_{auto} + LearnGC$.
The average Dice of scar is $ 0.702 \pm 0.071 $, which is comparable to the inter-observer variation ($ 0.695 \pm 0.049 $), and the difference is not significant ($p$=0.7783).
This conclusion also applies when we compare them using accuracy and \emph{G}Dice evaluation metrics.

\begin{figure}[tb]\center
 \includegraphics[width=0.46\textwidth]{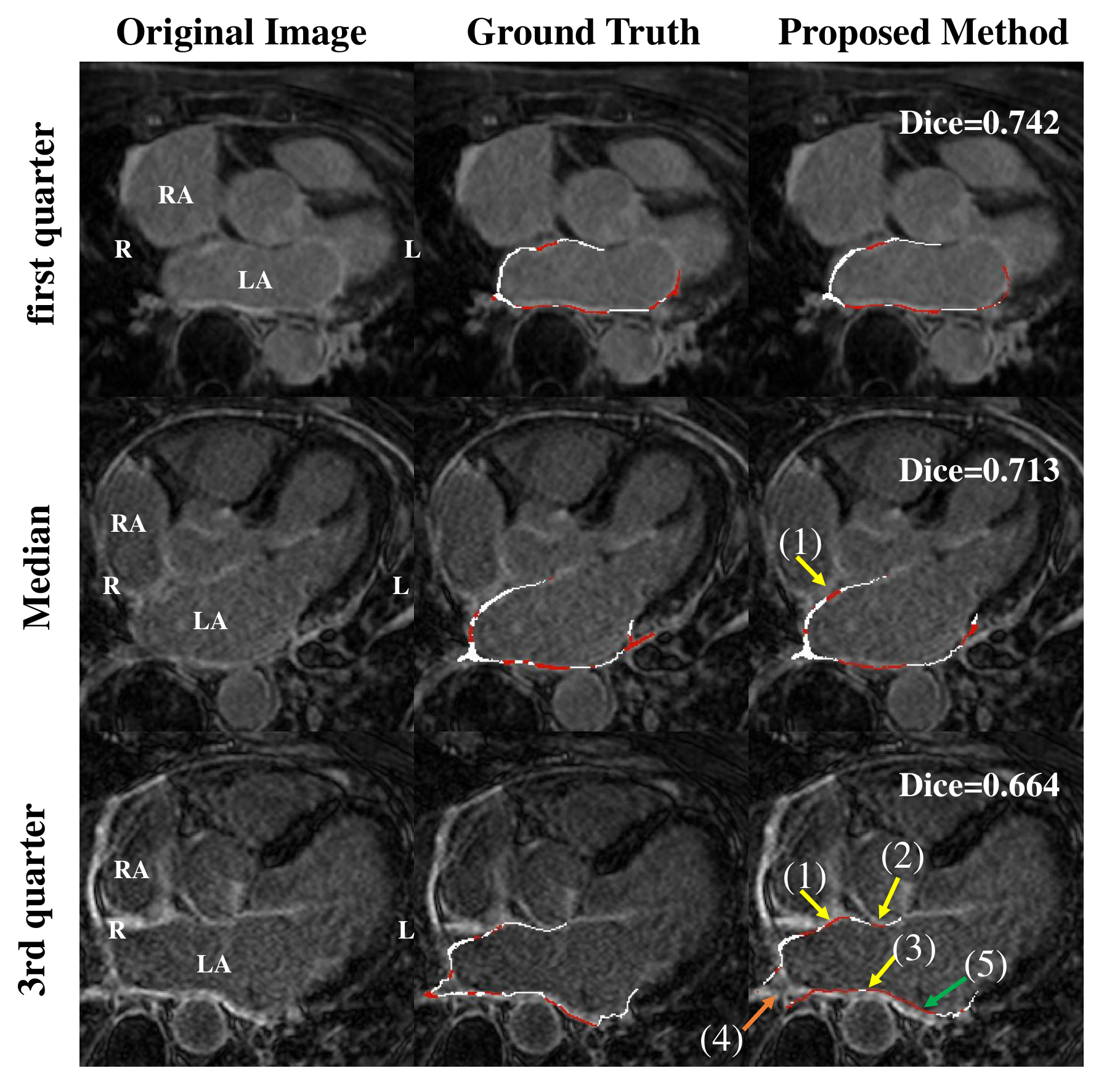} 
   \caption{Axial view of the images, the ground truth scar segmentation and the results by the proposed method. The red and white color labels represent the scar and normal wall, respectively. Arrow (1), (2) and (3) indicate the major classification errors of the proposed method caused by the surrounding enhanced regions, respectively from the right atrium wall, ascending aorta wall and descending aorta wall; arrow (4) shows an error from the misalignment between the automatic LA segmentation and the ground truth; arrow (5) illustrates that the proposed method can still perform well, even the automatic LA segmentation contains obvious errors.}
\label{fig:result:2dvisual}\end{figure}

\Zxhreffig{fig:result:2dvisual} provides 2D visualization of the axial view from three examples.
These three cases were the first quarter, median and third quarter cases from the test set in terms of Dice (scar) by the proposed method.
This illustrates that the method could provide promising performance for localizing and quantifying atrial scars of LA.
In the median and third quarter cases, we highlight the errors, particularly due to the enhanced adjacent regions, pointed out by arrow (1), (2) and (3).
These mis-classifications, representing the main challenges of this task, contributed to the major errors of scar quantification by the proposed method.
Another type of error was caused by the misalignments of the automatic LA segmentation, as arrow (4) pointed out.
This happened in some local areas where the errors occurred because of the different shapes of LA after reconstruction from the automatic segmentation.
One can also see that even there existed large LA segmentation errors, indicated by arrow (5) in \zxhreffig{fig:result:2dvisual}, the proposed method still could identify the scars at the corresponding location of the projected surface.
This is mainly attributed to the effective training of the MS-CNN, which assigns random shifts along the perpendicular direction of the surface when extracting the training patches.
The multi-scale learning also contributes to the less demanding of accuracy from the automatic LA segmentation, thus enables to achieve fully automated LA scar quantification.

\newcommand{\tabincell}[2]{\begin{tabular}{@{}#1@{}}#2\end{tabular}}
\begin{table*} [t] \center
    \caption{
    Overview of previous methods for scar quantification and segmentation in LA. Abbreviations: segmentation (seg); inter-observer variation in terms of Dice (Inter-ob); Society of Photo-Optical Instrumentation Engineers (SPIE), IEEE Journal of Translational Engineering in Health and Medicine (TEHM), IEEE transactions on medical imaging (TMI), Medical physics (MP), Medical Image Computing and Computer-Assisted Intervention (MICCAI).
     }
\label{tb:table:review}
{\small
\begin{tabular}{ l| l *{5}{@{\ \,} l }}\hline
Work     &  No. subjects&  LA (wall) seg & Scar seg method & Result (Dice) & \qquad Inter-ob\\
\hline
\citet{conf/mi/Perry2012}, SPIE         \qquad &34&    manual     &  k-means                   & $0.807 \pm 0.106$  & \qquad $ 0.786 \pm 0.072 $\\
\citet{journal/jtehe/Karim2014}, TEHM   \qquad &15&    semi-auto  &  GMM + Graph-cuts          & $>0.8$             & \qquad N/A \\
\citet{journal/tmi/Ravanelli2014}, TMI  \qquad &10&    semi-auto  &  NVI + Manual correction   & $0.850 \pm 0.070$  & \qquad N/A \\
                                               &10&    auto       &  NVI                       & $0.600 \pm 0.210$  & \qquad N/A \\
\citet{conf/miccai/wu2018}, MICCAI      \qquad &36&    auto       &  Multivariate mixture model& $0.556 \pm 0.187$  & \qquad N/A \\
\citet{conf/miccai/Chen2018}, MICCAI    \qquad &100&   auto       &  Dilated Attention Network & $0.776 \pm 0.146$  & \qquad N/A \\
\citet{journal/mp/Yang2018}, MP         \qquad &37&    auto  &  Super-pixels + SVM        & $0.790 \pm 0.050$  & \qquad N/A \\
\hline
\end{tabular} }
\end{table*}

\section{Discussion and conclusion} \label{conclusion}
In this work, we have proposed a fully automatic framework for segmentation and quantification of LA scars.
Two major methodological contributions have been introduced.
One is the formulation of quantifying the LA scarring based on a surface mesh.
The classification and quantification are achieved via the surface projection and graph-cuts framework.
The other is the adoption of the multi-scale learning combined with CNN, i.e. MS-CNN.
The multi-scale learning is implemented using the MSP strategy, which extracts the features from both the local and global intensity profiles of LGE MRI.
The MS-CNN learns both the label probability of each nodes and the relations between connected nodes in the graph.
The surface projection in the proposed framework avoids the difficulty of providing an accurate and demanding LA wall segmentation,
and the multi-scale patch-based learning, with the random shift training strategy, further mitigates the effect of less accurate LA initialization from a fully automatic approach, as demonstrated in Section \ref{exp_randomshift}.
We employed fifty-eight images with manual delineation for experiments.
The proposed method performs better when the size of extracted patches increases, but the performance converges when the size is larger than a certain value (see Section \ref{exp_patchsize}).
The multi-scaling learning further improves the performance compared to the method with single-scale learning, as demonstrated in Section \ref{exp_scale}.
Finally, the proposed learning graph-cuts based method demonstrates evidently better performance compared to the conventional approaches,
and the mean accuracy and Dice (scar) for quantifying LA scars are respectively 0.856 and 0.702, which are comparable to those of inter-observer variation (accuracy=0.867, Dice =0.695).

\Zxhreftb{tb:table:review} summarizes the related works from literature.
\citet{conf/mi/Perry2012} evaluated their method on a dataset consisting of 34 images.
The mean Dice score was $ 0.807 \pm 0.106 $, and the inter-observation Dice was $ 0.786 \pm 0.072 $.
Their method required an accurate initialization of LA walls from manual segmentation, followed by a k-mean classification.
\citet{journal/jtehe/Karim2014} employed GMM to model the enhancement of scar region, and used the graph-cuts method to consider neighbouring regions.
This method used LA segmentation for initialization, which was achieved from a semi-automatic method with manual correction.
They evaluated the method using numerical phantoms as well as using 15 {\it in vivo} images. They obtained more than 0.8 Dice scores on the two datasets.
\citet{journal/tmi/Ravanelli2014} adopted a threshold based approach,
where the normalized voxel intensity (NVI) of LA walls was applied.
The threshold value, $NVI=4$, was assigned according to previous studies and visual validation by experts,
base on which they used a 2-D skeletonization algorithm to quantify the atrial fibrosis.
The authors evaluated both the fully automatic method and the semi-automatic approach with manual correction.
The mean Dice scores of LA scar quantification increased from $ 0.60 \pm 0.21 $ to $ 0.85 \pm 0.07 $ when the manual correction was included.
\citet{conf/miccai/wu2018} proposed a fully automatic method for LA fibrosis quantification.
They formulated the joint distribution of images based on the multivariate mixture model, and optimized model parameters using the iterated conditional mode algorithm.
They tested the method on 36 cases and reported a mean Dice score of $0.556 \pm 0.187$ and average accuracy of $0.809 \pm 0.150$.
\citet{conf/miccai/Chen2018} developed a multi-view two-task recursive attention model for simultaneous segmentation of LA and scars.
The mean Dice score of LA segmentation was $0.908 \pm 0.031$, which was similar to the result (Dice=$ 0.898 \pm 0.044 $) from our study, though their average Dice score of scar quantification was $0.776 \pm 0.146$.
\citet{journal/mp/Yang2018} employed the super-pixel algorithm and SVM to segment the scars on 37 subjects.
They obtained $0.790 \pm 0.050$ Dice score, 0.87 segmentation accuracy, 0.89 sensitivity and 0.79 specificity by using the leave-one-out cross-validation strategy.
This study yielded better Dice score than ours in this work, but there was no evident difference in terms of the accuracy, sensitivity and specificity between these two works.
It should be noted that among these six works, only one, i.e., \citet{conf/mi/Perry2012}, reported the details of inter-observer variation.
Also note that it can be difficult to pursue an objective cross-study comparison due to the difference of datasets, initialization methods, and evaluation metrics.

One of the challenges of LA scar quantification is to distinguish artifacts from the boundary regions, such as from the RA wall and aorta wall, as we discussed above and showed in \zxhreffig{fig:challenges} and \zxhreffig{fig:result:2dvisual}.
Conventionally, providing accurate LA walls is the crucial step \citep{journal/jcmr/Karim2013,conf/mi/Perry2012}.
In this work, we propose to use multi-scale deep learning technology, with specifically designed training strategy, to tackle this challenge.
However, due to the limited training data, the errors caused by this problem could still happen.
Secondly, the quantification of scars in our work is performed on the surface mesh projected from the LA endocardium.
\citet{journal/mia/Karim2018} discussed the importance of wall thickness, particularly considering the potential that the ectopic activity can prevail in scars that are non-transmural.
However, they also emphasized that the relationship between the AF and the changes in wall thickness was not clear, and the thickness was difficult to measure based on current MRI data.
In clinical practice, the location and extent of scarring areas are considered to have greater clinical significance,
which is however arduous to represent and to perform quantitative cross-subject comparisons.
In the future work, visual assessment will be considered.
Finally, a limitation of this work is that the gold standard was constructed from the manual segmentation of only one cardiologist. In the future, we can combine the delineations from multiple experts to obtain an average and consensus gold standard.


\section*{Acknowledgement}
This work was supported by the Science and Technology Commission of Shanghai Municipality (17JC1401600) and the NSFC (81301283).
This study was also funded by the British Heart Foundation Project Grant (Project Number: PG/16/78/32402).

\section*{References}
\bibliographystyle{model2-names}
\bibliography{AllBibliography_lei_new}

\end{document}